\documentclass{llncs}

\usepackage{amsfonts}
\usepackage{subfigure}
\usepackage{amsmath} 
\usepackage{amssymb} 
\usepackage{amsbsy} 
\usepackage{amstext}
\usepackage{harpoon}
\usepackage{array}
\usepackage{latexsym}
\usepackage[T1]{fontenc}
\usepackage[english]{fp-autoref}
\def\sectiontext{Section}
\def\figuretext{Figure}

\usepackage{algorithm}
\usepackage[noend]{algpseudocode}

\makeatletter
\def\BState{\State\hskip-\ALG@thistlm}
\makeatother

\usepackage{booktabs}
\usepackage{tabularx}
\usepackage{rotating}
\usepackage{multirow}
\usepackage{array}
\usepackage{colortbl}
\usepackage[table]{xcolor}

\newcommand{\noun}[1]{\textsc{#1}}

\usepackage{fp-frame}
\usepackage[latin1]{inputenc}
\usepackage{mathpartir}
\usepackage{moreverb}
\usepackage{stmaryrd}
\usepackage{url}
\usepackage{epsfig}
\usepackage{alltt}
\usepackage{mathrsfs} 
\usepackage{bbm}  
\usepackage{wasysym} 
\usepackage{protosem}
\usepackage{setspace}
\usepackage{ifthen}
\usepackage{boxedminipage} 
\usepackage{listings}
\usepackage{color}
\PassOptionsToPackage{usenames,dvipsnames}{xcolor} 
\usepackage{fancyvrb}
\usepackage{graphics}
\usepackage{tabularx}
\usepackage{graphicx}
\usepackage{cite}
\usepackage{url}
\usepackage{hyperref}

\hypersetup{
    colorlinks=true,
    citecolor=blue,
    linkbordercolor=black,
    urlcolor=blue,
    linkcolor=blue}

\usepackage{tikz}
\usepackage{booktabs}
\usetikzlibrary{calc}

\newcommand{\tikzmark}[1]{\tikz[overlay,remember picture] \node (#1) {};}
\newcommand{\DrawBox}[3][]{%
    \tikz[overlay,remember picture]{
    \draw[black,#1]
      ($(#2)+(-1.05em,3.4ex)$) rectangle
      ($(#3)+(1.85em,-1.5ex)$);}
}

\newcommand{\tikzmarksmall}[1]{\tikz[overlay,remember picture] \node (#1) {};}
\newcommand{\DrawBoxsmall}[3][]{%
    \tikz[overlay,remember picture]{
    \draw[black,#1]
      ($(#2)+(-0.7em,3.4ex)$) rectangle
      ($(#3)+(0.3em,-1.5ex)$);}
}

\newcommand{\allerrors}
{
\begin{table*}
	\centering
	\def\arraystretch{1.4}
	\definecolor{lightgray}{gray}{0.95}
	\rowcolors{3}{lightgray}{}
	\renewcommand{\tabcolsep}{0.065cm}
    \caption{\label{table:allerrors}
    Median mean-absolute error
    with corresponding standard errors in parentheses.
    Only the best testing filter- and wrapper-based results (choice of $R$) are displayed.
    We explicitly compare GPESA with the state-of-art, SL.
    Bold values indicate significance (at 0.05 level with Bonferroni correction)
    under a Wilcoxon singed rank test 
    in which the null hypothesis asserts that distribution of the
	differences between GPESA and SL
	is symmetrically distributed about 0.
	}
    \scriptsize
    \begin{tabular}{ccccccccccccccccccc}
	\toprule
	\textbf{Year} & & \noun{SR} & & \tikzmarksmall{top left 2}\noun{SL} & & \noun{SGP} & & \noun{FR$_4$} & & \noun{FL$_{19}$} & & \noun{FGP$_{19}$} & & \noun{WR$_2$} & & \noun{WL$_3$}  & & \tikzmark{top left 1} \noun{GPESA}  \\
	\midrule
	2003 && 0.86 && 0.51 && 0.35 (0.14) && 0.50 && 0.46 && 0.44 (0.08) && 0.43 (0.10) && 0.49 (0.09) && \textbf{0.29 (0.09)} \\
	2004 && 0.47 && 0.30 && 0.32 (0.10) && 0.34 && 0.29 && 0.26 (0.05) && 0.37 (0.16) && 0.35 (0.16) && \textbf{0.17 (0.05)} \\
	2005 && 0.95 && 0.44 && 0.50 (0.13) && 0.61 && 0.40 && 0.52 (0.06) && 0.58 (0.11) && 0.63 (0.09) && \textbf{0.32 (0.07)} \\
	2006 && 0.66 && 0.27 && 0.41 (0.29) && 0.57 && 0.52 && 0.36 (0.06) && 0.53 (0.11) && 0.54 (0.11) && 0.27 (0.05) \\
	2007 && 0.72 && 0.33 && 0.44 (0.10) && 0.42 && 0.38 && 0.34 (0.05) && 0.52 (0.13) && 0.50 (0.11) && \textbf{0.24 (0.06)} \\
	2008 && 1.46 && 0.46 && 0.60 (0.13) && 0.71 && 0.64 && 0.58 (0.11) && 0.70 (0.31) && 0.54 (0.26) && 0.52 (0.18) \\        
	2009 && 0.81 && 0.41 && 0.65 (0.08) && 0.90 && 0.61 && 0.56 (0.08) && 0.98 (0.10) && 1.03 (0.09) && 0.41 (0.10) \\
	2010 && 0.62 && 0.48 && 0.44 (0.12) && 0.43 && 0.47 && 0.41 (0.06) && 0.43 (0.11) && 0.52 (0.11) && \textbf{0.32 (0.07)} \\
	2011 && 0.87 && 0.48 && 0.61 (0.17) && 0.77 && 0.60 && 0.53 (0.10) && 0.82 (0.20) && 0.93 (0.16) && 0.45 (0.12) \\ \midrule
    \rowcolor{white} Mean && 0.82 && 0.41 \tikzmarksmall{bottom right 2}&&	0.48        && 0.58 && 0.49 && 0.44        && 0.58        && 0.61        &&	0.33 \tikzmark{bottom right 1}\\
	\bottomrule
	\end{tabular}
    \DrawBox[thick, draw=blue, dotted]{top left 1}{bottom right 1}
    \DrawBoxsmall[thick, draw=blue, dotted]{top left 2}{bottom right 2}
\end{table*}
}

\newcommand{\heatmaps}
{
\begin{figure}
\centering
\includegraphics[width=1\textwidth]{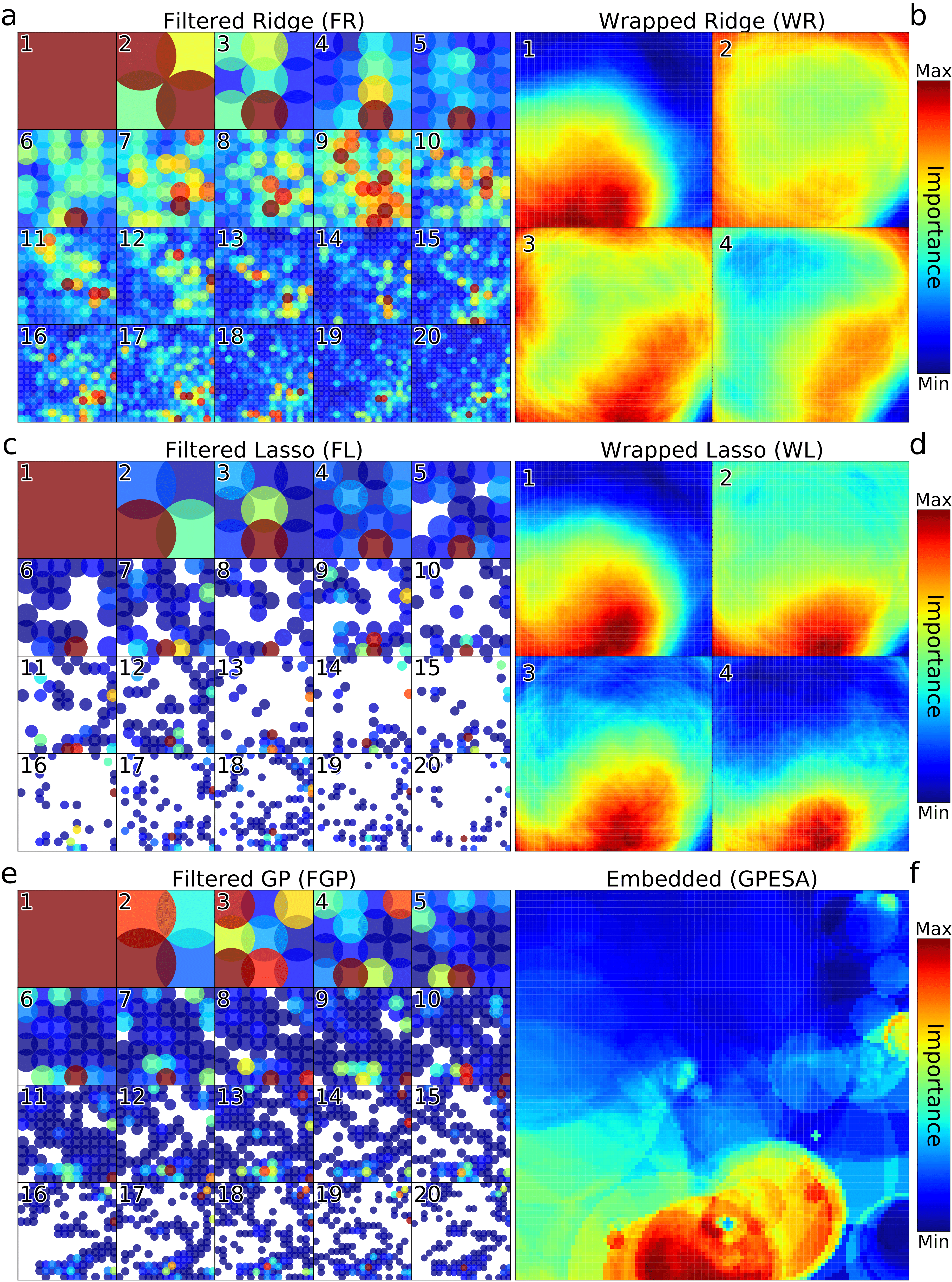}
\caption{\label{figure:heatmaps} 
		Importance (defined in Section \ref{section-discussion}) of
        spatial units. For filters a.) FR, c.) FL, and e.) FGP, importance is
        displayed at each resolution $R \in \{1, 2, \dots 20\}$
     	and each individual filter subplot is annotated with
        the corresponding $R$. For wrappers b.) WR and d.) WL, 
        $R \in \{1, 2, 3, 4\}$. Finally, f.) GPESA, which 
        has no $R$ parameter. White areas indicate spatial units unused 
        in feature construction across all three exploratory variables. 
        }
\end{figure}
}

\newcommand{\summary}
{
\begin{figure}
\centering
\includegraphics[width=1\textwidth]{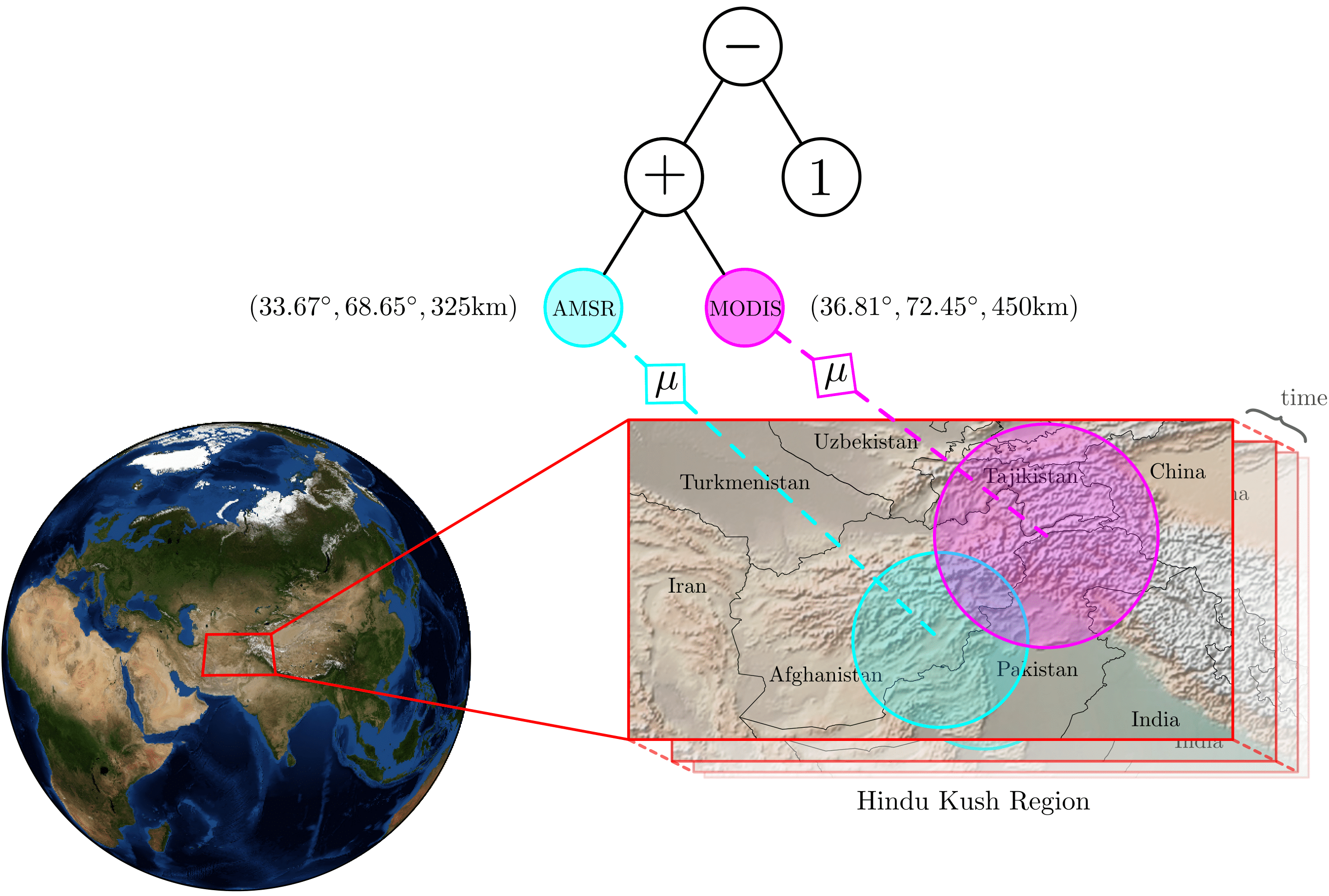}
\caption{\label{figure:summary}
		In GPESA, standard GP terminals are replaced with specialized aggregation operators that allow for the automation of the number, location and size of aggregations, and the way in which the they are combined to form a predictive model. Here, two such aggregations, taken from the evolved cyan and magenta regions (specified by their latitude, longitude and radius), are combined and adjusted by a scalar value.
        }
\end{figure}
}

\renewcommand{\paragraph}[1]{\textbf{\textit{#1.}}}

\begin{document}

\title{Evolving Spatially Aggregated Features From Satellite Imagery for Regional Modeling}
%
\titlerunning{Evolving Aggregated Features}  
%

\author{Sam Kriegman \and Marcin Szubert \and Josh C Bongard \and Christian Skalka}
\authorrunning{Sam Kriegman et al.} 
%
\tocauthor{Sam Kriegman, Marcin Szubert, Josh C Bongard, and Christian Skalka}
\institute{University of Vermont, Burlington VT 05405, USA,\\
\email{sam.kriegman@uvm.edu}
}

\maketitle

\begin{abstract} 

Satellite imagery and remote sensing provide explanatory variables at
relatively high resolutions for modeling geospatial phenomena, yet
regional summaries are often desirable for analysis and actionable
insight. 
In this paper, we propose a novel method of inducing spatial aggregations
as a component of the machine learning process, yielding
regional model features whose construction is driven by model prediction
performance rather than prior assumptions.
Our results demonstrate that
Genetic Programming is particularly well suited 
to this type of feature construction
because it can automatically synthesize appropriate aggregations, as well
as better incorporate them into predictive models compared 
to other regression methods we tested.
In our experiments we consider a specific problem instance and
real-world dataset relevant to predicting snow properties in
high-mountain Asia.


\end{abstract}

%
\section{Introduction}
\label{section-intro}

Regional modeling focuses on explaining phenomena 
occurring at a regional, as opposed to site-specific or global scales\cite{rees2009regional}.
Regional models 
are of interest in many remote sensing applications,
as they provide meaningful units for analysis and actionable insight
to policymakers. Yet satellite imagery and remote sensing provide
variables at relatively high resolutions.  Consequently, studies often
involve decisions concerning how to integrate this information in
order to model regional processes. 
Considering measurements at each individual spatial unit 
as a separate model feature
can result in a high dimensional problem in which high variance and
overfitting are major concerns.
For this reason, spatial aggregation is
often applied in this setting to uniformly up-sample variables to be
consistent with the response.  
Although in averaging variables across all spatial units in
the region, we discard information
which could in turn diminish prediction accuracy and our
understanding of underlying phenomena.

Rather than strictly incorporating individual spatial units or 
uniformly up-sampling, 
it might instead be beneficial to construct features of a regional 
model using particularly important subsets of geographical space.
In this
paper, we move away from uniform up-sampling aggregations towards more
flexible and interesting aggregation operations predicated on their
subsequent use as features of a regional model. We propose a novel 
method of inducing spatial aggregations 
as a component of the machine learning process, 
yielding features whose construction is driven by model
performance rather than prior assumptions.  

In experiments designed to explore these techniques, we
consider a specific problem and real dataset: estimating 
regional Snow Water
Equivalent (SWE) in high-mountain Asia with 
satellite imagery. 
Improved estimation of SWE in
mountainous regions is critical \cite{Dong05} but is difficult
due in part to complex characteristics of snow
distribution \cite{buckingham-skalka-bongard-jhydro15not}.

\section{Methods}
\label{section-methods}

We take a comparative approach to the SWE problem, considering ridge
regression, lasso, and GP-based symbolic regression\footnote{The source code necessary for reproducing our results is available at \newline
\url{https://github.com/skriegman/ppsn_2016}.}.  For each
regression model, we consider a filter-based method of feature
construction in addition to a second, more dynamic method.  For linear
regression, we incorporate a wrapper approach in which constructed
features and the regression model are induced in separate learning
processes, with feedback between the two. For symbolic regression, we
use an embedded approach where constructed features and the regression 
model are induced simultaneously over the course of an evolutionary run.

\subsubsection{The Dataset.} The SWE dataset\footnote{Raw satellite data was 
	pre-processed by Dr.~Jeff Dozier (UCSB) using
	previously reported techniques and is available upon request.} 
is derived from data collected by NASA's Advanced Microwave
Scanning Radiometer (AMSR2/E) and Moderate Resolution Imaging
Spectroradiometer (MODIS) for March 1 - September 30, in 2003 - 2011, 
over an area that spans most of the high mountain Asia.
We have three explanatory variables
measured daily across a $113 \times 113$ regular grid for 1935 days:
(1) mean and (2) standard deviation of sub-pixel Snow Covered Area \cite{Painter09, Dozier08},
as well as (3) an estimate of SWE derived from passive microwaves \cite{tedesco2010assessment}.
Our response variable is regional SWE, an attribute of the
entire study region, represented as a single value
for each of the 1935 days. 
The response was ``reconstructed'' by combining snow cover
depletion record with a calculation of the melt rate to retroactively
estimate how much snow had existed in the region
\cite{martinec1981areal}.

\subsection{Regression Models}
\label{section-methods-algorithms}

Ridge regression \cite{hoerl1970ridge} is similar to 
ordinary least squares (OLS)
but subject to a bound on the $L_2$-norm of the coefficients. 
Because of the nature of its quadratic constraint, ridge
regression cannot produce coefficients exactly equal to zero
and keeps all of the features in its model.
Lasso (Least Absolute Shrinkage and Selection Operator,
\cite{tibshirani1996regression}) modifies the ridge 
penalty and is subject to a bound on the $L_1$-norm of the coefficients.
The geometry of this $L_1$-penalty has a strong tendency 
to produce sparse solutions with coefficients exactly equal to zero.
In many high dimensional settings, lasso is the state-of-the-art
regression method given its
ability to produce parsimonious models
with excellent generalization performance.
For both lasso and ridge regression, the parameter 
constraining the coefficients is set through cross-validation.

Genetic Programming
(GP, \cite{Koza_1992_Genetic}) is a very flexible heuristic technique
which can conveniently represent free-form mathematical equations
(candidate regression models) as parse trees. 
GP's inherent flexibility is well-suited for our particular
problem because it can efficiently 
express spatial aggregations and seamlessly
combine them into the learning process with minimal assumptions.
Furthermore, the ``white box'' nature of GP may provide
physical insights about this complex problem that is currently
lacking, as in other domains \cite{Bongard07, Schmidt11b}.

To search the space of possible GP trees we use a variant of
Age-Fitness Pareto Optimization (AFPO, \cite{Schmidt_2011_Age-Fitness}).
AFPO is a multiobjective method that relies on the concept of
genotypic age, an attribute 
intended to preserve diversity.  
We extend AFPO to include an additional objective of model size,
defined as the syntactic length of an individual tree. 
The size attribute protects
parsimonious models which are less prone to overfitting the 
training data.
The GP algorithm therefore 
identifies the Pareto front
using three objectives (all minimized):
age, error (fitness), and size. For the fitness objective, we use a
correlation-based function rather than pure error, and define
$ f_{COR} = 1 - |\phi(\hat{s}, s)|, $
where $\phi(\hat{s}-s)$ denotes Pearson correlation between
model predictions 
($\hat{s}$) 
and actual values of our response 
($s$), 
regional SWE. 
Correlation has recently been shown to outperform
error-based search drivers 
given that if
a model makes a systematic error 
it could be easily eliminated 
by linearly scaling the output and therefore should be protected \cite{stanislawska2015genetic}. 
Accordingly, for all GP implementations, 
we apply a linear transformation after $f_{COR}$ -driven 
evolution has concluded, by using an individual program (model) 
output as the single input of OLS 
on the training data. 

Our implemented GP experiments used ramped half-and-half
initialization with a height range of $2-6$ and  
an instruction set including unary ($\{\sin, \cos, \log,$ $ \exp\}$) 
and binary functions ($\{\times, +, -, /\}$). 
One thousand
individuals in the
population are subject to crossover (with probability 0.75) and mutation
(with probability 0.01) over
the course of 1000 generations. There is a static limit on the tree
height (17) as well as the tree size (300 nodes).  Each experiment
consists of 30 evolutionary runs, from which the best model
(lowest training $f_{COR}$) is selected.
The selected model is then transformed using OLS, 
and subsequently validated using unseen test data.

\paragraph{Standard Methods} Ridge regression, lasso, and GP may be performed on the raw data
using each variable at each individual spatial unit as a separate feature.
We denote these methods as Standard Ridge (SR), Standard Lasso (SL) and 
Standard GP (SGP).
SR, SL and SGP each have access to $113 \times 113 \times 3=38307$ features, but only 1720 
observations in each fold of data. 

\subsection{Feature Construction Methods}
\label{section-feature-construction}

Feature construction is a well studied problem
and the utility of 
genetic programming for feature construction 
has been recognized in many previous studies \cite{krawiec2002genetic}.
The key difference in our work from this past work 
is the nature of the data being modeled. 
We presume that there exist spatial autocorrelations
of varying size and shape
that, if aggregated to improve the signal to
noise ratio, yield features supporting more accurate
predictions.

In a regional model, we can construct features by
aggregating higher dimensional variables across space.
However,
it is not entirely clear what kind of aggregations 
are useful as features of a predictive model. Grouping variables based on similarity or dissimilarity 
does not necessarily produce useful regional features.
In this paper, we make an assumption about the importance of distance 
and continuity in effective spatial aggregations, based on 
Tobler's first law of geography \cite{tobler1970computer} 
which states that
``everything is related to everything else, 
but near things are more related than distant things.''
Accordingly, we limit the space of possible spatial aggregations 
to be an average of values within a circular spatial area 
defined by its centerpoint and radius. 
However, where to aggregate, how many aggregations
to perform, and how to combine the aggregates must
still be determined manually or decided during model
optimization.
We view filters and wrappers as intermediary 
steps
in relaxing assumptions towards our embedded approach,
which automates all three of these
aspects.

\subsubsection*{The Filter Method.}
\label{section-experiments-filter}

Filter-based feature construction methods transform or ``filter'' the
original variables as a preprocessing step, prior to modeling.  Our
filter for the SWE problem represents a static up-sampling
transformation of the original variables.  Each variable is decomposed
in space by a grid of overlapping circles\footnote{The shape of
  circles are in reality so-called ``small circles,'' as they lie on
  the surface of earth.}  of equal radii centered on a square lattice
pattern of points (see Figure \ref{figure:heatmaps}a,c,e for example).  
Each constructed feature corresponds to the average
(arithmetic mean)
of a particular variable sampled within a particular circle of space.
Units that reside in an overlapping region of two separate circles are
included in the calculation of both features.  
Since there are three
explanatory variables in the SWE dataset, an $R \times R$ grid
corresponds to $p = 3R^2$ constructed features.  The
constructed features are then used as inputs for ridge regression, lasso, and GP,
which we will refer to as Filtered Ridge (FR), Filtered Lasso (FL),
and Filtered GP (FGP). We will also specify
the value of $R$ used in a particular model instance as 
a subscript, e.g.~FR$_{15}$ denotes
Filtered Ridge with $R$=15.
We consider filters with $R \in \{1, 2, \dots, 20\}$, however
note that the standard methods are essentially filters with $R=113$,
albeit with the non-overlapping square pixels.

\subsubsection*{The Wrapper Method.}
\label{section-experiments-wrapper}

Wrapper-based feature construction methods incorporate feedback from
the fit of the model. We implement wrappers around 
both ridge regression and lasso in order to 
enable the circular sampling regions to define their own
center and radius. The circles are no longer fixed on a grid
with a predetermined size.
Instead, each constructed feature
is uniquely parameterized by the coordinates of a center unit $(x,
y)$, as a latitude and longitude tuple, and a radius $r$, as a single
value floating point number in km.  The center can be any spatial unit
in the region, including one at the edge of
the raster.  The radius is restricted to be within 0 and 1000 km,
which is flexible enough to contain only a single unit or span the
entire region (see Figure \ref{figure:heatmaps}b,d for example).

Wrapped Ridge (WR) and Wrapped Lasso (WL) separately 
use a ridge/lasso-driven hill climbing algorithm
to construct features that
minimize Mean Absolute Error (MAE), i.e.
$ \frac{1}{n} \sum_{i=1}^n |\hat{s_i}-s_i|, $
where $s_i$ is the actual value of our response (regional SWE)
and $\hat{s_i}$ is output predicted by the model over $n$ observations.
The algorithm uses the same number of circles for 
each of the three variables, initializing
their parameters $(x, y, r)$ randomly.
For 1000 iterations, a single constructed feature (circle)
is randomly selected and subject to a Gaussian 
mutation on one of its parameters with standard deviation equal to 
25\% of the radius and centered at zero. A new ridge/lasso model is then
refit on the mutated set of features using a random subset of data
sampled without replacement.
If the mutation lowered model error on the complementing
set of training data left out, then the change is 
accepted. Otherwise, the mutation is undone.
If a proposed mutation to the radius would take it outside the
restricted range of $0-1000$ km, 
then it is ``bounced-back'' the distance it
would have exceeded the boundary. For example, a random mutation
that would result in a radius of 1200 km, 
becomes $1000-(1200-1000)=800$ km.
Thirty restarts are used from which the best model 
based on training data is selected.	
We consider $R \in \{1, 2, 3, 4\}$ for wrappers corresponding to $3 \times R^2$
features which really means $3 \times 3 \times R^2$ modifiable parameters.

\subsubsection*{The Embedded Method.}
\label{section-experiments-embedded-gp}
By using GP, we can allow for flexibility with respect to the placement and number of aggregations as well as the way in which they are combined to form a model.
However,
stochastic optimization methods like GP cannot be 
easily ``refit'' in the same manner as deterministic
algorithms like ridge regression or lasso. 
Therefore using wrapper approach for GP is computationally infeasible.
Instead, modifications to aggregated features 
are implemented through mutation-based operators. 

In Genetic Programming with Embedded Spatial Aggregation (GPESA)
introduced here, our constructed features 
are represented as parameterized 
tree terminals, with parameters $(x, y, r)$ (Figure \ref{figure:summary}).
Constructed features are
randomly initialized in the same manner as the wrapper method,
but separately for each terminal of each individual in the population.
Greedy Gaussian mutations to the parameters $(x, y, r)$
of a randomly selected constructed feature occur in the population
with 20\% probability, each generation.
Mutations to $r$ have mean zero and a standard deviation 
of 25\%, subject to the bounce-back rule.
Similarly,
mutations to $(x,y)$ have mean distance zero and a 
standard deviation of $0.25r$.
For 25 iterations,
greedy mutations modify the parameterized terminals 
within a particular GP tree. A modification is accepted if
it successfully reduces average error ($f_{COR}$) 
on random subsets of training data sampled with replacement.
Aside from the 
stochastic application, 
another key difference between the wrapper method's hill climbing
algorithm and the GPESA's greedy mutations is that 
the overall regression model stays the same between mutations 
rather than being ``refit'' after each mutation.

\subsubsection{Validation.}
\label{section-methods-validation}

In order to validate the generalization of models we
partition the dataset into nine overlapping folds. 
Each fold corresponds to leaving out
one year for testing and training on the remaining eight (using years
2003 - 2011). We use MAE on the unseen test data as a metric to assess model performance.
To account for a difference in scale across any set of features, 
all input model features are standardized over time
by removing the mean and scaling to unit variance.
This means that as wrapper and embedded methods construct new
aggregations, 
the sampled data is scaled over time prior to being averaged over space.
Since our goal is near-real-time estimation for a future day,
the training values of a feature's mean and variance
are reapplied when scaling the same feature in validation.

\summary

\section{Results}
\label{section-results}

Table \ref{table:allerrors} displays the test error
of each valid regression 
and feature construction method
combination. 
For filters and wrappers, 
only the best performing model is displayed
and we indicate the particular value of parameter $R$ as a subscript.
Since the ultimate goal of our paper is to synthesize 
a method better than existing approaches,
we must statistically compare 
GPESA to SL, the state-of-the-art linear regression / variable
selection algorithm.
The null hypothesis of interest here is that of no difference between GPESA and a SL.
Therefore we perform yearly Wilcoxon signed rank tests \cite{hollander2013nonparametric}
comparing GPESA to SL
with Bonferroni correction across the nine years. 
For five out of the nine test years, GPESA
is significantly better than SL,
while for the other four years there is no significant difference 
with SL.

\allerrors

Through displaying only the best testing filters and wrappers, 
we aim to focus speculation about GPESA performance
through a conservative lens.
Yet we ultimately 
view filters and wrappers as intermediary 
steps ``working up'' to GPESA.
Accordingly, the best test error better represents  
a bound on the potential performance of a particular intermediary method
even though it may not be possible to achieve such performance through
a parameter sweep based on the training data.
And indeed, across all methods tested,
GPESA reported the lowest recorded median mean-absolute error within all but two 
years (7 of 9) where it has the second lowest.

\section{Discussion}
\label{section-discussion}

Our results show that incorporating dynamic aggregations of 
higher resolution variables into a
regional model is beneficial in our particular problem setting, as
compared to both uniform up-sampling of variables and a state-of-the-art
linear regression technique (SL) that incorporates individual spatial units.  SL achieves
competitive prediction performance through a sparse linear combination
of the individual spatial units, on par with SGP which is not linearly
constrained.
Ultimately, GPESA performed significantly better 
(lower median test error) than
SL on a majority (5 of 9) of cross validation folds.
Moreover, whenever GPESA was not significantly better than SL it was not significantly worse.

A main reason why GPESA has an advantage in this application is 
the difficulty of knowing a priori what the most 
important spatial datapoints are, 
and how to best aggregate them.
Additionally, 
the structure of the model itself is unknown and it depends on 
the resulting aggregations.
Therefore this is not a fixed length optimization problem,
which makes it well-suited for 
GPESA, which can search over different
numbers and non-linear combinations
of spatial aggregations.
While SL can theoretically perform the same
aggregation as a
GPESA terminal (mean within a radius of a geographical point),
SL is restricted to a single linear solution while GPESA is not.

However, it's important to emphasize that the computational cost
of GPESA is higher than that of traditional GP 
and much higher than that of linear regression. 
In particular, the most expensive operation is 
the ``on the fly'' aggregation component of GPESA 
which makes the fitness evaluation require 500\% more time than in SGP.
Part of the incurred cost is due to inefficiencies 
of our implementation that necessitated a copy 
with all spatial aggregation operations. 
In future work we will look at reducing 
this overhead through more efficient data structures (e.g. k-d trees).

\subsubsection{Importance of Spatial Data.} 
To better understand the relevance of particular spatial locations, we
define the \emph{importance} of a spatial unit for both linear and
symbolic methods, separately.  For ridge regression and lasso, we can
define importance by exploiting the disposition of coefficients to be
larger for variables with a stronger correlation to the response,
relative to a particular feature set.  We define linear regression
importance of a particular spatial unit as the average absolute
coefficient of features that incorporate the unit into a regression
model. 
While we cannot as easily
determine relative importance within nonlinear models, we can instead
define importance by exploiting the multiple candidate solutions
provided from stochastic multiobjective optimization.  
We define GP importance of a particular spatial unit as the
average absolute correlation ($1-f_{COR}$) of nondominated solutions
that incorporate the unit.

To visualize the importance of spatial information, we generated a
series of heatmaps (Figure \ref{figure:heatmaps}).
In Figures \ref{figure:heatmaps}a,
\ref{figure:heatmaps}c, and \ref{figure:heatmaps}e we show regional
importance values of filter methods for each $R \in \{1,...,20 \}$,
with the relevant value of $R$ annotated in the upper left corner of
each box. Note that in lasso- and GP-based approaches, some variables
are unused (white), while ridge cannot perform variable selection and
uses all.  Figures
\ref{figure:heatmaps}b and \ref{figure:heatmaps}d plot WR and WL for
$R \in \{1,2,3,4\}$.  Finally, Figures \ref{figure:heatmaps}e and
\ref{figure:heatmaps}f plot the importance of spatial information in
the GP sense, for FGP and GPESA, respectively.  Overall, this
visualization indicates an agreement among all methods on the
relatively higher importance of information in the lower center/right
region of the image.

\heatmaps

\section{Conclusion}
\label{section-conclusion}

In this work we developed a novel method to address the problem of
modeling a regional response with high resolution satellite imagery.
We moved away from uniform up-sampling aggregations towards more
flexible and interesting aggregation operations predicated on their
subsequent use as features of a regional model.  Our proposed
technique, GPESA, is general and intended to apply to a variety of
modeling problems on spatially organized data. But as an application
example, and as a setting in which to evaluate our techniques, we
considered the problem of estimating snow water equivalent in high
mountain Asia using satellite imagery.  Our results showed that using
GP to evolve spatial aggregations outperforms lasso, the
state-of-the-art method for directly incorporating individual spatial
units into a sparse linear model.

In future work we plan to explore more flexible spatial and temporal
aggregations for more predictive modeling in real earth science applications. 

\medskip

\noindent\textbf{Acknowledgements:} Thanks to Dr.~Jeff Dozier (UCSB)
for posing the high-mountain Asia SWE problem and providing
associated datasets. 


\bibliographystyle{abbrv}
{
\bibliography{references,main}


\end{document}